# Text recognition on images using pre-trained CNN

**Afgani Fajar Rizky[1], Novanto Yudistira[1], Edy Santoso[1]**
[1]Department of Informatics Engineering, Faculty of Computer Science, University of Brawijaya, Malang, Indonesia

## ABSTRACT

A text on an image often stores important information and directly carries high level semantics, makes it as important source of information and become a very active research topic. Many studies have shown that the use of CNN-based neural networks is quite effective and accurate for image classification which is the basis of text recognition. It can also be more enhanced by using transfer learning from pre-trained model trained on ImageNet dataset as an initial weight. In this research, the recognition is trained by using Chars74K dataset and the best model results then tested on some samples of IIIT-5K-Dataset. The research results showed that the best accuracy is the model that trained using VGG-16 architecture applied with image transformation of rotation 15°, image scale of 0.9, and the application of gaussian blur effect. The research model has an accuracy of 97.94% for validation data, 98.16% for test topic, and 95.62% for the test data from IIIT-5K-Dataset. Based on these results, it can be concluded that pre-trained CNN can produce good accuracy for text recognition, and the model architecture that used in this study can be used as reference material in the development of text detection systems in the future.

**Corresponding Author:**

Afgani Fajar Rizky
Intelligent System Laboratory, Faculty of Computer Science, University of Brawijaya
Veteran, 65145, Malang, East Java, Indonesia
Email: avajar@student.ub.ac.id

## 1. INTRODUCTION

A text on an image often stores important information, such as street names, vehicle license plates, and someone's personal information. As a product of human abstraction and manipulation, text in natural scenes directly carries high level semantics. This property makes text present in natural images and videos a special, important source of information [1]. For that reason, scene text detection has become a very active research topic in recent several years [2]. As the rise of deep learning, scene text detection has reaches state-of-art performances [20]-[22], however, its performance on text tecognition still has to be extensively validated. Thus, this research was conducted to get a better result in text recognition.

Convolutional Neural Network (CNN) is a machine learning algorithm that is often used in image recognition. CNN has an excellent performance in machine learning problems, especially the applications that deal with image data [3]. Since the success of CNN in recognizing digital images on dataset called MNIST [23], CNN is continued to be applied on several application. On challenging big natural image dataset, deep CNN achieved state-of-the-art performance leaving the traditional handcrafted features with machine learning behind [24]. For individual character recognition study of the 'ICDAR 2003 Character Database' dataset which contains character images with blur and distortion [25], CNN algorithm has been used in which reaches an accuracy rate of 84.53% [4]. A similar study regarding text recognition in images but with a different method, was carried out by applying optical character recognition with a neural network that achieved an accuracy of 97.58% [5]. Thus, the CNN algorithm was chosen as the method in this study.

In improving the performance of the model, the transfer learning method is commonly used as the initial weight initiation in the CNN algorithm. The use of transfer learning in the character dataset showcases the ability of the model to learn and adapt to a target dataset with limited training samples [6]. The use of transfer learning has also been carried out on handwritten text recognition and has shown good performance when trained in small databases [7].

Based on the problems above, this research was carried out with the aim to find the best model for text recognition by using transfer learning on CNN classification. The result then will be tested to find the model with the best accuracy value.

## 2. METHOD

In this study, the text recognition is proposed by using classification of character images. The classification result will be used as a training model that will be tested on every cropped character on scene text image. The method or research flow used in this study is divided into 2 parts: character recognition part, and scene text processing part. Each part is divided into several stages, in the character recognition part, the flow consists of: character data collection, data loading and augmentation, model training, and model evaluation. In the scene text processing part, the flow consists of: scene text data collection, image processing, bounding box detection and cropping, and model evaluation on scene text data. The research flow is shown in Figure 1(a) for character recognition, and Figure 1(b) for scene text processing.

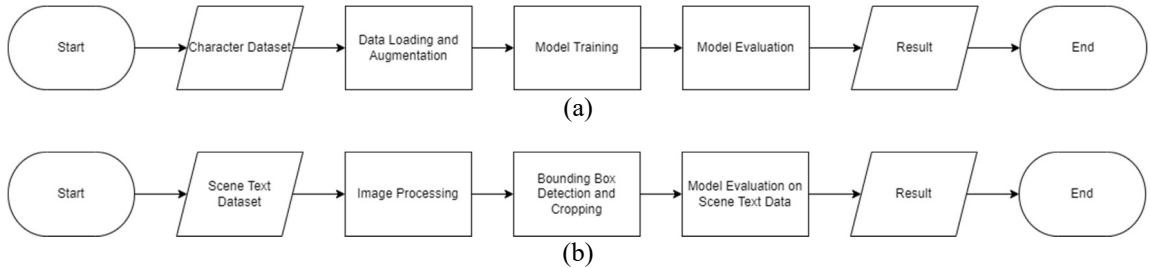

Figure 1. Research flow for (a) character recognition and (b) scene text processing

In the character recognition part, we use alphabet characters data as training and evaluation data, the data is loaded into a data loader and applied by augmentation effect to add more data variety and normalize the data. The training data is then trained by using CNN algorithm to get a recognition model with a good result. Model generated by the training process is then evaluated by using validation and test data. This evaluation result is then analyzed to get the model with the best accuracy. In the scene text processing part, we use scene text data as an evaluation data from the model that has been trained earlier. Firstly, the scene text data is being processed by using image processing techniques to help the bounding box detection process easier. After the image is processed, we detect the bounding box of each character by calculating each contour in the image. This bounding box is then cropped and saved as a new image, which then used as test data and used as evaluation for the model earlier. The evaluation result is analyzed to calculate the accuracy of the model.

### 2.1. Dataset

The dataset used in this study consisted of two types of data, which is cropped character image data for each alphabetic character, and scene text data that is captured in the outdoor environment. The dataset used is the Chars74K dataset [8] and the IIIT 5K-word dataset [9].

### 2.1.1. Chars74K

The Chars74K dataset is a dataset that contains character images consisting of three types of images: handwritten images, computer font images, and scene text images. The Chars74K dataset focuses on recognizing characters in situations that would traditionally not be handled well by OCR techniques [8]. The Chars74K dataset consists of two types of image data: alphabetic images and Kannada images. In this study, the data used is an alphabetic image, which contains 36 labels with a total of 26 letter characters (a-z) and 10 numeric characters (0-9). The sample data from the Chars74K dataset is shown in Figure 2.

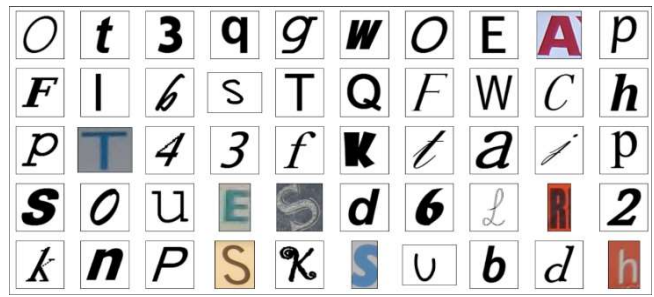

Figure 2. Chars74K sample data

### 2.1.2. IIIT 5K-word

The IIIT 5K-word dataset is a dataset obtained based on Google Image search results. Data is collected using various query words such as billboards, signboards, and various other queries. This dataset contains 5000 pieces of word text images taken from external environmental images and digital images. The IIIT 5K-word dataset was released with a focus on the problem of recognizing text extracted from natural scene images and the web [9]. In this study, the final model chosen after the training and testing process will be tested on 30 sample data taken from IIIT 5K-word dataset. The sample data choosen is shown in Figure 3.

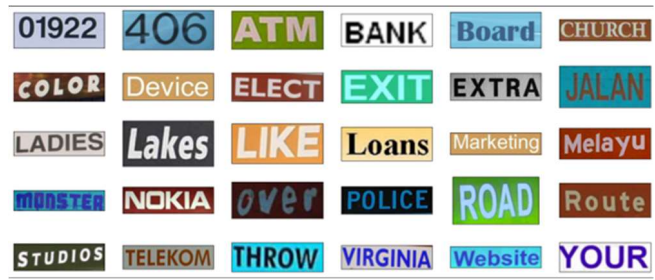

Figure 3. IIIT 5K-word sample data

### 2.2. Image processing

Image processing is an image manipulation process aimed at improving image quality by reducing noise and sharpening certain image features [10]. Another goal of image processing is to get a cleaner and lighter image to make it easier for further processing. Image processing can be done in many ways, such as color processing, shape processing, etc. We use image processing to get a bounding box in each character on scene text image, each bounding box detected then cropped and saved into a new image, which then used as test data. The image processing steps used in this research are shown in Figure 4.

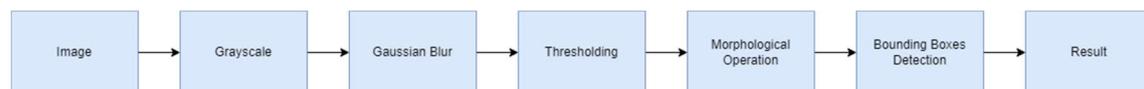

Figure 4. Image Processing Process

The first step in the image processing process is by manipulating the image color by converting it into grayscale format. By converting it, we change the color scale from RGB value into single value by eliminating the color information and leaving only luminance information on the image on a scale of 0-255. The next step is to apply gaussian blur into the image to remove noise and irrelevant information. The amount of gaussian blur can be adjusted according to the size of the kernel used. The larger the kernel used, the less noise is obtained, but the information lost will also be greater. After converting the image into grayscale and applying blur, we can use image thresholding to change image color into binary scale (black and white) to make bounding box detection easier. Image thresholding is done by setting a limit value, if the color scale value is smaller than the limit value, then the color will turn into black, and vice versa. The setting can also be reversed which is usually referred to as "inverse binary thresholding". After that, we can apply a morphological effect to enlarge the black or white area in the image. Morphological operation can be divided into dilation, erosion, opening, and closing. The one we used in this study is opening to sharpen the white area in the image while retaining the original size to help bounding box detection more easily. Lastly, we can apply and draw the bounding box based on white area in the image from the calculated contour. The example result of image processing steps is shown in Figure 5.

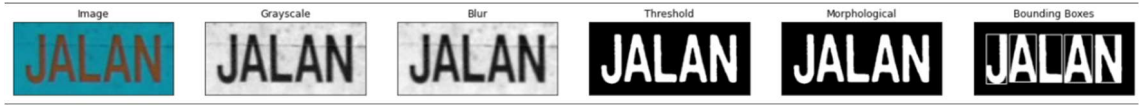

Figure 5. Example Result

## 2.3. Convolutional neural network

CNN or Convolutional Neural Network is a type of deep learning neural network algorithm that is often used in the image classification process. Recently, deep learning algorithms have had great success in various computer vision problems [11]. In its application, CNN implements a convolutional layer which is used to carry out the convolution process by creating a kernel based on the data section of the input data. The output of the convolutional layer is then processed at the pooling layer which is used to reduce the amount of data without losing important features by taking data that has been determined by the formula on certain pieces of data, such as the average value or the maximum value. This process is then repeated with an amount as the total of convolutional and pooling layers. Then, the data is generally flattened to convert the data into one long dimension. And for the last step, the data is then going into the fully-connected layer for the classification process to get the output of the correct label from the CNN process. In this study, the vanilla CNN architecture used for the experiment is illustrated in Figure 6.

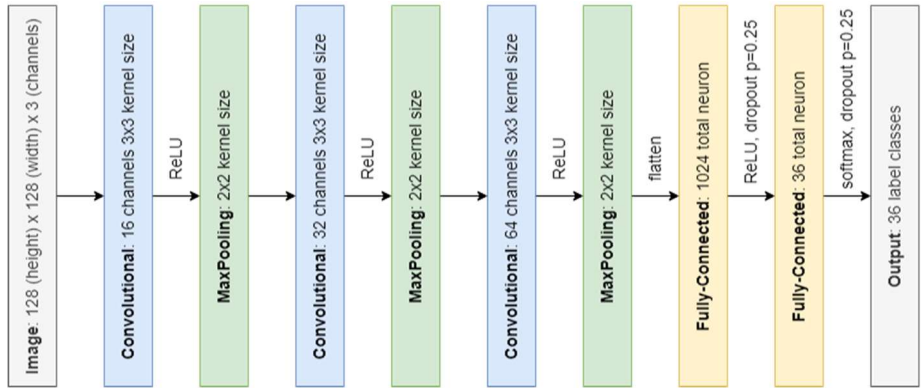

Figure 6. Illustration of CNN architecture used as vanilla model

### 2.3.1. ReLU function

The ReLU (Rectified Linear Unit) activation function is a non-linear activation function that is used to eliminate neurons that have negative output results, and are replaced with a value of 0. Thus, the ReLU activation function is computationally efficient since it will deactivate the neuron with values below 0. The formula of ReLU function is expressed in Equation 1.

$$R(x) = max(0, x)) \tag{1}$$

Where $R(x)$ is ReLU result and $x$ is neuron value.

### 2.3.2. Softmax function

Softmax Activation Function is a non-linear activation function used in categorical classification. Softmax is widely used to map outputs of neural networks into a categorical probabilistic distribution for classification [12]. The probability distribution is used as the probability of the data to the data class. The softmax activation function is calculated by dividing the sigmoid value of a value by the sum of all the sigmoid values contained in the vector, so that the resulting probability vector has a total value of 1. Softmax functions are expressed in Equation 2.

$$\sigma(z)_i = \frac{e(z_i)}{\sum_{j=1}^{K} e(z_j)}) \tag{2}$$

Where $\sigma$ is softmax result, $(z)_i$ is input vector, $e$ is exponential function through applied element, $K$ is number of classes, $z_i$ is element of input vector, and $z_j$ is an element of vectors in all classes.

### 2.3.3. Cross entropy loss

Loss function is a function that is used to calculate the difference between the predicted results of a model and the actual value. In the case of classification, Cross entropy is a suitable loss function by calculating the sum of the output log values on the target label. The cross entropy function are expressed in Equation 3.

$$L = \frac{-1}{N} \sum_{i=1}^{N} y_i \, log(\hat{y}_i) \qquad (3)$$

Where $L$ is loss value, $N$ is total data, $y_i$ is target label, and $\hat{y}_i$ is output value.

### 2.3.4. SGD optimization

Optimization in deep learning is a method of updating the weights by evaluating the performance of the model against the loss function. Stochastic Gradient Descent (SGD) is an optimization algorithm for updating the weights by getting the gradient loss value against the layer weight value by considering the amount of learning rate. Stochastic gradient descent (SGD) in contrast performs a parameter update for each training example x(i) and label y [13]. The SGD optimization are expressed in Equation 4.

$$W_{t+1} = W_t - \alpha \frac{dL}{dW_t} \qquad (4)$$

Where $W_{t+1}$ is new layer weight, $W_t$ is old layer weight, $\alpha$ is learning rate, $dL$ is derivative value of loss result, and $dW_t$ is the derivative value of old weight.

### 2.4. Transfer learning

Transfer learning is a method that works by using the knowledge that has been obtained by a deep learning model, known as a pre-trained model, to solve a problem, and then this knowledge will be used to solve a new problem [14]. Practically, transfer learning is the process of using a pre-trained model that has been trained on a large dataset against a new model with the aim of getting the features that have been trained on the dataset and as the initiation of weights on the new model. Transfer learning is generally divided into 2 parts, features extraction and classifier. In this experiment, features extraction will keep using the same architecture as pre-trained model architecture, while the classifier will be tuned and use the same architecture as the vanilla model for all of the pre-trained models.

### 2.4.1. AlexNet model

AlexNet is a deep learning architecture model that won the ImageNet Large Scale Visual Recognition Challenge (ILSVRC) competition on September 30, 2012. The ImageNet architecture has a total of 8 layers consisting of 5 convolution layers and 3 fully-connected layers. The AlexNet architecture is well known for some of its innovations at the neural network layer, such as ReLU functionality, dropout layers, and overlapping pooling layers. The AlexNet model contains a number of new and unusual features which improve its performance and reduce its training time [15]. An illustration of the AlexNet architecture used in this experiment is shown in Figure 7.

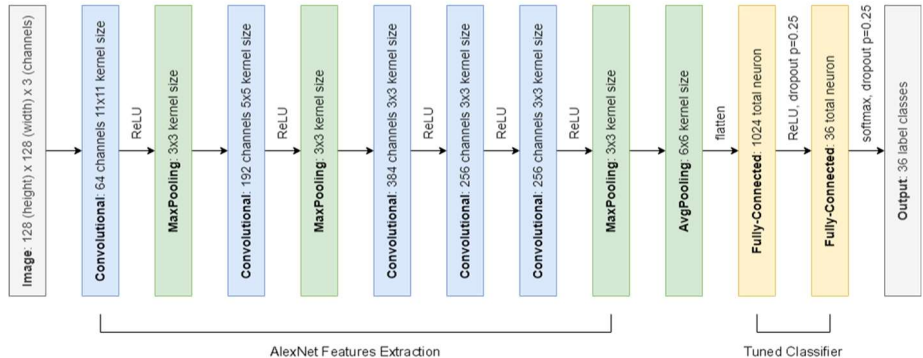

Figure 7. Illustration of AlexNet model architecture used

### 2.4.2. VGG-16 model

The VGG model is a CNN architectural model that was a runner up in ILSVRC competition in 2014 by a group called the Visual Geometry Group from Oxford University. The VGG architecture was proposed with the aim of discussing another important aspect of ConvNet architecture design – its depth [16]. The VGG architecture is unique in using multiple convolution layers to increase layer depth by using several small 3x3 convolution filters that replace the large kernel filters in the AlexNet architecture. An illustration of the VGG architecture is shown in Figure 8.

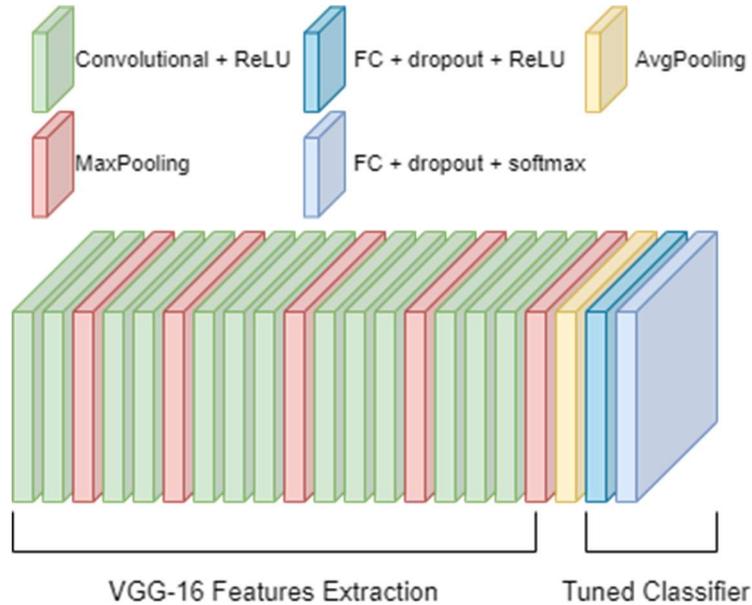

Figure 8. Illustration of VGG16 model architecture used

### 2.4.3. ResNet-18 model

The ResNet architectural model is an architectural model that was introduced in 2015 and won 1st place in the ILSVRC-2015 competition. The ResNet model architecture is introduced by proposing the concept of a skip connection mechanism with the aim of solving the vanishing gradient problem. The vanishing gradient problem is a problem where the gradient value of the loss function can reach a value of 0 when the number of networks used is too deep. With the network depth increasing, accuracy gets saturated [17]. This causes the process of updating the weights in the initial layer cannot occur because the gradient value reaches 0 and the weight cannot be changed. An illustration of the ResNet architecture is shown in Figure 9.

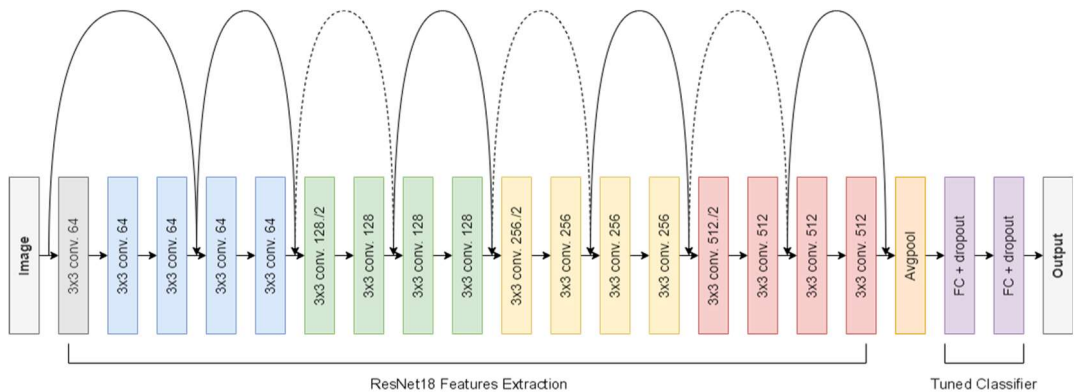

Figure 9. Illustration of ResNet18 model architecture used

### 2.4.4. DenseNet-121 model

The DenseNet model is an architectural model that was introduced at the CVPR conference in 2017 and was awarded the 'Best Paper Award' at the conference. The DenseNet architectural model introduces the concept of dense blocks by connecting each layer of the neural network directly to each other. To ensure maximum information flow between layers in the network, we connect all layers (with matching feature-map sizes) directly with each other [18]. An illustration of the DenseNet architecture is shown in Figure 10.

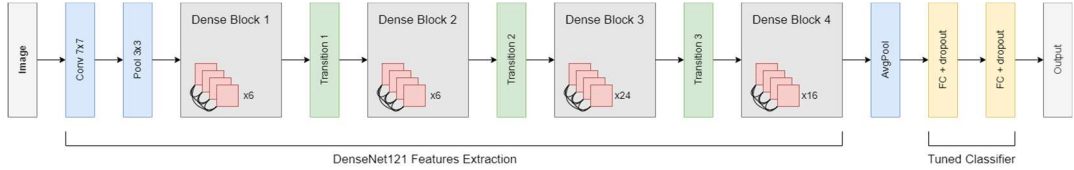

Figure 10. Illustration of DenseNet121 model architecture used

### 2.5. Freeze layer

Freeze layer is a method in neural network to freeze the initial layer of the model so that gradient calculations and weight updates do not occur in the initial layer of the model. This is intended to reduce computation time, but still maintain model accuracy. This is done by training only certain parts of the entire layer, progressively "freezing" the layer and removing it from the backward pass process [19].

## 3.   RESULTS AND DISCUSSION

In this section, it discussed about the training configuration that used in every experiment and about the experiment result & the discussion from its result.

### 3.1.  Experiment configuration

In the training technique process, there are several predefined hyperparameters that are used for all experiments equally, so that there is no bias in the results caused by differences in hyperparameters. In these experiments, the hyperparameters that used includes such as the total number of batch size of 64, total training epochs of 100 epochs, image size of 128x128 pixels, learning rate size of 0.01, loss function of cross entropy loss, and optimizer algorithm of stochastic gradient descent. In addition, the default augmentation for training and test set is resize, convert to tensor, and normalize. The experimentation was also configured using Google Colaboratory GPU for its training and testing process.

### 3.2.  Discussion

This session will explain the experiment results and the explanation along with it. The experiment is divided into four sections to discuss the experiment results for the effect on using image augmentation, transfer learning, and freeze layer for model accuracy, and also for the overall accuracy of the selected model based on previous experiment in scene text image.

### 3.2.1. Image augmentation impact on model accuracy

The experiment on image augmentation is conducted to find the ideal parameter values for each type of augmentation type that is used in training data. Augmentation is used to increase image variation in the training data with the aim that the model can adapt better to various image conditions at the test data. The augmentation used is adjusted to the type of image that will be transformed. In the character image case, the augmentation that will be used are image random rotation, image random scale, and image random effect. The experimentation results for each augmentation are shown in Table 1 for image random rotation, Table 2 for image random scale, and Table 3 for image random effect.

Table 1. Experiment results of random rotation augmentation

| Total Degree | Training Accuracy | Validation Accuracy | Test Accuracy |
|---|---|---|---|
| 0° | 99.44 | 94.80 | 95.11 |
| **15°** | **98.97** | **94.83** | **95.61** |
| 30° | 98.76 | 94.67 | 95.41 |
| 45° | 98.46 | 94.11 | 94.54 |
| 60° | 98.18 | 93.63 | 94.73 |

Table 2. Experiment results of random scale augmentation

| Scale Size | Training Accuracy | Validation Accuracy | Test Accuracy |
|---|---|---|---|
| 1.0 (Default) | 98.97 | 94.83 | 95.61 |
| 0.9-1.0 | 98.93 | 94.81 | 95.03 |
| **0.8-1.0** | **98.65** | **95.05** | **95.50** |
| 0.75-1.0 | 98.45 | 94.91 | 95.14 |
| 0.5-1.0 | 98.09 | 94.74 | 95.09 |
| 0.5-0.75 | 98.19 | 66.45 | 65.73 |

Table 3. Experiment results of random effect augmentation

| Effect | Training Accuracy | Validation Accuracy | Test Accuracy |
|---|---|---|---|
| Default | 98.65 | 95.05 | 95.50 |
| **Blur** | **98.71** | **94.97** | **95.63** |
| Grayscale | 98.61 | 94.52 | 94.98 |
| Blur + Grayscale | 98.64 | 94.58 | 94.98 |

The highest accuracy values for each augmentation are found in the augmentation process using a rotation of 15°, an image scale of 0.9, and the use of a blur effect on the image. The final results of the highest accuracy value are 98.71% for training data, 94.97% for validation data, and 95.63% for test data.

### 3.2.2. Transfer learning impact on model accuracy

The experiment is conducted to test the effect of pretrained weights on the architectural model that had been trained with the ImageNet dataset. Pretrained weights that have been trained have been shown to provide better results for test results. The architectural models used in this research are AlexNet architecture, VGG-16 architecture, ResNet-18 architecture, and DenseNet-121 architecture. The experiment results are shown in Table 4.

Table 4. Experiment results of transfer learning model

| Model | Training Accuracy | Validation Accuracy | Test Accuracy |
|---|---|---|---|
| vanilla model | 98.71 | 94.97 | 95.63 |
| AlexNet | 99.13 | 97.45 | 97.50 |
| **VGG16** | **99.43** | **97.94** | **98.16** |
| ResNet18 | 99.85 | 97.72 | 97.83 |
| DenseNet121 | 99.85 | 97.87 | 97.91 |

The highest accuracy value on validation data and test data was obtained using the VGG-16 model. The highest accuracy value was obtained at the 100th epoch training. In the epoch training, the accuracy value of the training data was obtained with a value of 99.43%, then the accuracy value of the validation data was obtained with a value of 97.94%, and the accuracy value of the test data was 98.16%. The loss value in the validation data is 0.0134 and the test data is 0.1092. So, the VGG-16 architecture will be used in the next experimentation. The graph of accuracy and loss value for each epoch is shown in Figure 7.

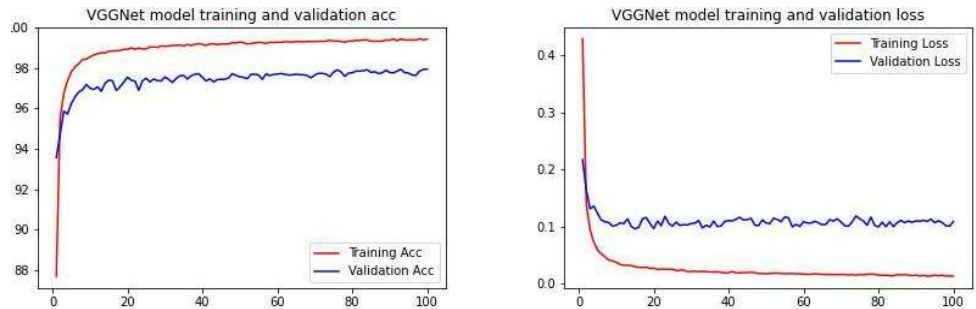

Figure 11. Accuracy and loss graph of VGG-16 model

### 3.2.3. Freeze layer impact on model accuracy

The experiment is conducted to test the effect of using freeze layer in model accuracy. In several previous studies, there are some conclusions that using freeze layer in the initial layer of architecture, can reduce the training time significantly, while keeping up the model performance. So, in this study, freeze layer experimentation was carried out to obtain the ideal number of layers that got frozen. The experiment results are shown in Table 5.

Table 5. Experiment results of freeze layer total layer

| Model | Training Accuracy | Validation Accuracy | Test Accuracy |
|---|---|---|---|
| **0** | **99.43** | **97.94** | **98.16** |
| 1 | 99.42 | 97.93 | 98.02 |
| 2 | 99.35 | 97.90 | 98.13 |
| 3 | 99.38 | 97.87 | 98.13 |
| All Layers | 96.27 | 91.81 | 93.14 |

The highest accuracy value on validation data and test data is still found in the default model that didn't use any freeze layer. The use of a freeze layer also resulted in decreased accuracy in training and validation data until the use of 3 layers, experimenting to use a freeze layer on all layers also resulted in a drastic decrease in accuracy, so the test was stopped at 3 layers. Thus, the default model that didn't use any freeze layer will choosen as the final model for the next experiment.

### 3.2.4. Overall accuracy on scene text image data

The last experiment is to test the selected model to test the character in scene text data. The data used in this experimentation is 30 sample data taken from the IIIT-5K-word-dataset. The experiment is carried out by performing image processing to get the area of each character contained in the image dataset, which is then cropped into a new image that will be used as test data for the augmentation configuration, architectural model, and total freeze layer that has been selected. The experimentation result gives a total accuracy of 95.62%. The result of 5 samples used for this experiment is shown at Table 6.

Table 6. Sample of experiment results on scene text data

| Image | Actual Text | Predicted Text | Recognition Accuracy |
|---|---|---|---|
| 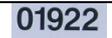 | 01922 | 0I922 | 4/5 (80%) |
| 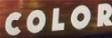 | COLOR | COLOR | 5/5 (100%) |
| 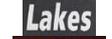 | LAKES | LAKES | 5/5 (100%) |
| 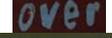 | OVER | OVET | 3/4 (75%) |
| 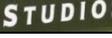 | STUDIOS | STUDIOS | 7/7 (100%) |

## 4. CONCLUSION

Based on the results of the experimentation process, it can be concluded that image augmentation on character dataset affects the accuracy results generated by the model by giving more variance of data. The best accuracy results are found in the augmentation process using a rotation of 15°, an image scale of 0.5, and the use of a blur effect on the image. The use of transfer learning also gives better accuracy results compared to the model that was created manually from scratch. The best accuracy results are found in the use of the VGG-16 model with an accuracy result of 97.94% for validation data, and 98.16% for test data. Using a freeze layer on earlier layers also didn't have any impact and give less accuracy result than the normal one. In the end, the model then used to test on scene image data, and gives accuracy of 95.62%.

## BIOGRAPHIES OF AUTHORS

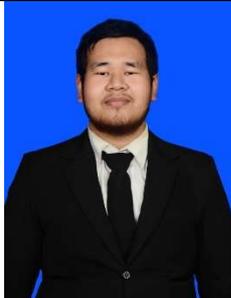 **Afgani Fajar Rizky**   is a student in progress of finishing computer science bachelor studies in University of Brawijaya that will be received in 2022. A young researcher that has a major interest in data science and machine learning related works. The research interest includes artificial intelligence, machine learning, neural networks, and computer vision. He can be contacted at email: avajar@student.ub.ac.id or afganifajarrizky@gmail.com.

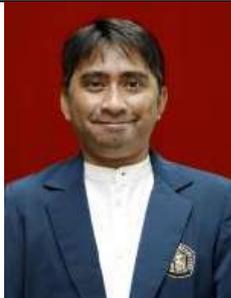 **Novanto Yudistira**     is currently lecturer and researcher at Brawijaya University, Indonesia. He received his BS in informatics engineering from the Institut Teknologi Sepuluh in November in 2007, his MS in computer science from Universiti Teknologi Malaysia in 2011, and his Dr. Eng. from information engineering, Hiroshima University, Japan in 2018. In 2016, he involved in research collaboration with Mathematical Neuroinformatics Group, National Institute of Advanced Industrial Science and Technology (AIST), Japan. In 2018, he continued Postdoctoral fellow in informatics and data science analytic for 2 years working with the Japanese large scientific research institute, RIKEN and Osaka university. His current research interests include deep learning, multi modal computer vision, medical informatics and big data analysis. He can be contacted at email: yudistira@ub.ac.id.

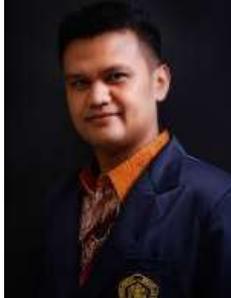 **Edy Santoso**    is a Lecturer and Researcher at the Department of Informatics Engineering, Faculty of Computer Science, Universitas Brawijaya. He completed his bachelor's degree in Mathematics Studies Program with an interest in Computer Science, Universitas Brawijaya. Then earned a Master's Degree at the Institut Sepuluh Nopember Surabaya in the field of Informatics Engineering. Various studies have been carried out especially in the fields of Artificial Intelligence. He can be contacted at email: edy144@ub.ac.id.